# MFMAM-YOLO: A Method for Detecting Pole-like Obstacles in Complex Environment

Lei Cai  Hao Wang*  Congling Zhou  Yongqiang Wang  Boyu Liu

*College of Mechanical Engineering, Tianjin University of Science & Technology, Tianjin 300222, China*

**Abstract**

In real-world traffic, there are various uncertainties and complexities in road and weather conditions. To solve the problem that the feature information of pole-like obstacles in complex environments is easily lost, resulting in low detection accuracy and low real-time performance, a multi-scale hybrid attention mechanism detection algorithm is proposed in this paper. First, the optimal transport function Monge-Kantorovich (MK) is incorporated not only to solve the problem of overlapping multiple prediction frames with optimal matching but also the MK function can be regularized to prevent model over-fitting; then, the features at different scales are up-sampled separately according to the optimized efficient multi-scale feature pyramid. Finally, the extraction of multi-scale feature space channel information is enhanced in complex environments based on the hybrid attention mechanism, which suppresses the irrelevant complex environment background information and focuses the feature information of pole-like obstacles. Meanwhile, this paper conducts real road test experiments in a variety of complex environments. The experimental results show that the detection precision, recall, and average precision of the method are 94.7%, 93.1%, and 97.4%, respectively, and the detection frame rate is 400 f/s. This research method can detect pole-like obstacles in a complex road environment in real time and accurately, which further promotes innovation and progress in the field of automatic driving.

## 1. Introduction

With the rapid development of computer vision technology and deep learning algorithms, obstacle detection has become an important technology in assisted driving and even autonomous driving systems [1-3]. Currently, deep learning-based obstacle detection methods have become mainstream, with commonly used architectures such as R-CNN [4], Fast R-CNN [5], Faster R-CNN [6-8], mask R-CNN [9], and YOLO [10]. At the same time, commonly used architectures have to be based on datasets. However, existing pole-like obstacle detection architectures face challenges such as a lack of underlying data sets and uneven quality of data sets, which limit the algorithm performance. In addition, the complex and variable traffic environment on urban roads [11], such as occlusion, rain or snow, etc. All of them can reduce the accuracy of pole-like obstacle detection. And due to the different heights and shapes of the pole-like obstacles, brings great challenges to the accuracy and real-time pole-like obstacle detection. At present, there are two main types of research on pole-like obstacle detection and classification. The first type of research is based on the LIDAR point cloud application. Plachetka C et al [12]. proposed a deep learning-based method for the detection and classification of pole-like objects in high-density point clouds. The precision of this method is 85%, but the real-time problem is not considered. Yanjun Wang et al [13]. proposed a method for extracting geometric and forest parameters of traveled road trees in onboard LIDAR point clouds for road scenes based on the Gaussian distributed region growth algorithm and Voronoi distance constraint. Although the method has a detection precision of 96.34% for the lane trees, the method consumes a large amount of storage space and computational resources resulting in a slow inference time. The second type of research is based on vision applications. Tengda Zhang et al [14]. proposed the OSLPNet model for extracting street light poles from street view images. The highest detection accuracy was 88.2% on a homemade data set, however, the detection speed was not considered. Sanjeewani P et al [15]. proposed a deep learning method based on single-class detection with 95.61% detection accuracy for curbside pole-like objects. Although the detection accuracy of all the current research methods to achieve more than 90%, the real-time is fully considered. Also, because the method is used for assisted driving environment perception system, both real-time and accuracy are required to be high. Therefore, based on the current complex and changing traffic environment on urban roads, this paper proposes a multi-scale feature hybrid attention module-look-only-once (MFMAM-YOLO). In summary, the main contributions of this paper are as follows:

1) The target detection loss function of the current YOLO framework mainly relies on the aggregation of the bounding box regression metrics, such as the distance, overlap area, and aspect ratio between the predicted and real boxes (i.e., GIoU [16], CIoU [17], DIoU [18], etc.), without taking into account the mismatch between the desired real box and the predicted box. This leads to the generation of multiple prediction frames and the failure to determine the best matching one in time when predicting multi-scale bounding boxes, which results in longer inference time and reduced detection accuracy. To solve the above problems, this paper improves the SIOU [19] loss function and introduces the optimal transport MK function [20]. The main idea is to transform the optimal matching problem that will predict the box into a linear programming problem and determine the optimal matching bounding box by an optimization algorithm. This reduces the inference of many bounding boxes and greatly speeds up the detection speed improving real-time performance and accuracy. At the same time, the MK function can regularize the model to prevent over-fitting [21] and name the improved loss function as MKS.

2) In such a complex road environment pole-like obstacles feature information is easily lost, while the size diversity of pole obstacles brings challenges to the extraction of feature information. Therefore, in this paper, based on the original pooling layer structure, SPP structure, and PAN structure, we use an efficient feature pyramid [22, 23] with a target detection algorithm for feature fusion. By introducing bottom-up and top-down cross-layer connections in the feature pyramid network, efficient extraction and fusion of feature information are achieved. This can effectively solve the problems of target multi-scale variation, target overlap, and target shape irregularity, with better robustness and generalization.

3) Due to factors such as mutual occlusion between pole-like obstacles and complex weather environments, various noises and darker light will occur when reading the image. To solve the above problems this paper embeds a CBAM [24] hybrid attention mechanism after the residual module in the backbone network of the MFMAM-YOLO model. Specifically, in the neck neural network, the CBAM module first performs the channel attention mechanism [25, 26] and then the spatial attention mechanism [26, 27] to enhance the useful channel and spatial information in the feature map. Thus, it helps the model to focus on important features when processing images and suppress irrelevant and complex background information to improve the accuracy and recall of the model. The remainder of this paper is organized as follows: Section 2 Introduced the method of target detection and classification. Section 3 presents the experimental results and the analysis of the results in detail. Finally, Section 4 concludes this paper.

## 2. Proposed Method

The YOLOv5 [28] network framework is currently the latest official object detection algorithm. Although unofficial single-stage object detection algorithms such as YOLOv6 [29]and YOLOv7 [30] have emerged, it does not necessarily mean that their detection performance is the best, as specific selection and improvement are needed depending on the target objects to be detected. The YOLOv5 model contains four types, YOLOv5s, YOLOv5m, YOLOv5l, and YOLOv5x, and its scale and number of parameters gradually increase. The rod obstacle detection system for assisted driving requires certain accuracy, real-time performance [31], and a lightweight model [32]. Although the YOLOv5s model is easy to deploy in assisted driving systems, the detection in complex road environments leads to low detection accuracy and real-time performance of existing algorithms. Therefore, this paper first uses the improved MKS loss function. Then, the[33-35] efficient fusion network architecture is fused with the cascaded hybrid attention mechanism to obtain the multi-scale feature hybrid attention module (MFMAM) for efficient extraction and fusion of features. Finally, the architecture of the MFMAM-YOLOv5s module proposed in this paper is shown in Fig. 1.

* Corresponding author.
  E-mail addresses: cailei0904@163.com (L. Cai), wangtn199011@tust.edu.cn (H. Wang), zhoucling@tust.edu.cn (C. Zhou), wangyq@tust.edu.cn (Y. Wang), lby19980923@163.com (B. Liu).

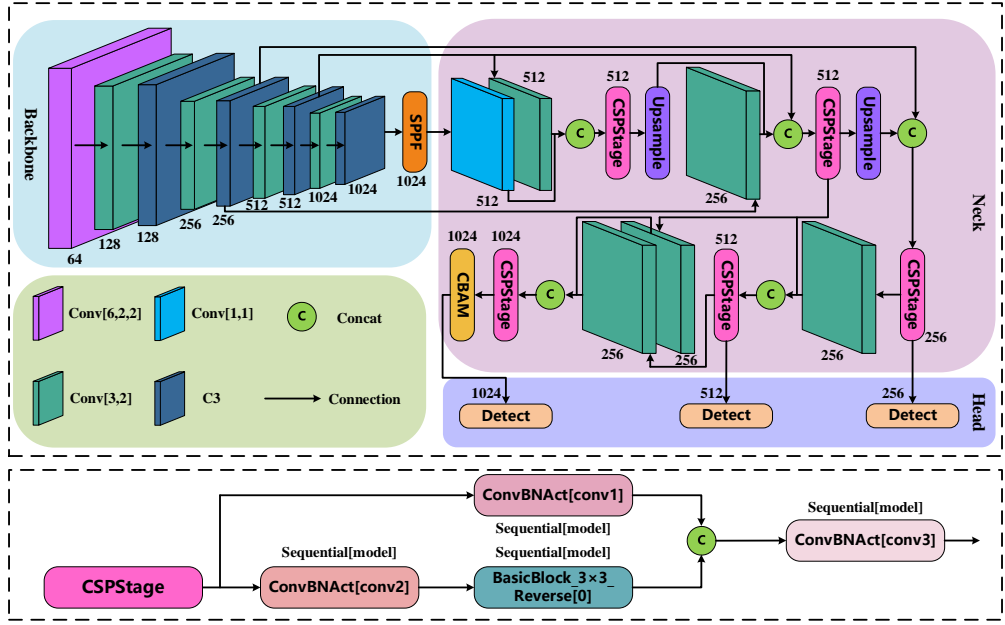

**Fig. 1** The architecture of the MFMAM-YOLOv5s network.

## 2.1. The Improved Loss Function

Object detection and recognition classification is a key research area in computer vision, and the loss function plays an important role in detection accuracy. Because the loss function is used to evaluate the difference between the prediction result of the model and the real target, the closer the prediction result is to the real target, the smaller the value of the loss function. the loss function of YOLOv5 mainly includes three aspects: bounding box loss (box loss), confidence loss (obj loss), and classification loss (cls loss) [36]. In the complex road environment, the model can only converge better if these three losses are minimized. However, there is no loss function specifically for pole-like obstacle detection. Therefore, in this paper, based on the SIoU loss function, the MK function is introduced to optimize the IoU loss function and redefine this loss function. Specifically, firstly, two sets of points are considered as input values i.e., the predicted bounding box and the true bounding box are considered as point sets. Then the distance matrix between these points is calculated, where each entry in the matrix represents the distance between the predicted bounding box and the true bounding box. Finally, the best match between the predicted bounding box and the true bounding box is found by the Sinkhorn-Knopp algorithm[37] to minimize the total negative IoU loss. In addition, the MK function can regularize the model to prevent over-fitting[38]. The MK function is then defined as follows:

The mathematical definition of the Monge problem[39] is: Given two metric spaces X and Y and two probability measures $\mu \in P(X), \nu \in P(Y)$, loss function $c : X \times Y \to R \cup \{+\infty\}$:

$$(MP) := \inf \left\{ \int_X c(x, T(x)) d\mu(x) : T_{\#\mu=\nu} \right\} \quad (1)$$

Where, c(x, T(x)) represents the loss to get from x to T(x), and $T_{\#\mu=\nu}$ represents the transport map between $\mu$ and $\nu$, i.e. T must be able to extrapolate the probability measure $\mu$ to $\nu$.

The Kantorovich problem[40] is an extension of the Monge problem, starting with the hope that the mound can be freely disassembled and freely combined. In the Monge problem, because the solution is a map, the mound can only be disassembled and cannot be freely combined. In the Kantorovich problem, two probability measures on two measure spaces X and Y are defined as probability measures $\gamma$ on X×Y spaces. The space of transport plans is defined as

$$\Pi(\mu, \nu) = \{\gamma \in P(X \times Y) \mid \pi_{X\#\gamma} = \mu, \pi_{Y\#\gamma} = \nu\} \quad (2)$$

The mathematical definition of the Kantorovich problem is that for two metric spaces X, Y, two probability measures $\mu \in P(X), \nu \in P(Y)$, loss function c:X×Y→R∪{+∞}:

$$(KP) := \inf \left\{ \int_{X \times Y} c(x, y) d\gamma(x, y) \mid \gamma \in \Pi(\mu, \nu) \right\} \quad (3)$$

To better understand this loss function, this paper redefines the relevant loss functions, including angle loss, distance loss, shape loss, and IoU loss. The angle loss is used to calculate the rotation angle of the detected frame. Since the rotation angle often requires a specific encoding, the angle loss can be used to ensure the correct angle encoding. Its definition is shown in Fig 2.

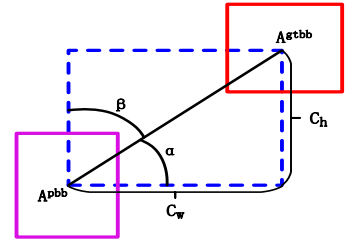

**Fig. 2** The relationship between the prediction box and the real box.

$$\wedge = \cos\left(2*\left(\arcsin(x) - \frac{\pi}{4}\right)\right) \quad (4)$$

$$x = \frac{c_h}{\sigma} = \sin(\alpha) \quad (5)$$

$$\sigma = \sqrt{\left(A_{c_x}^{gtbb} - A_{c_x}^{pbb}\right)^2 + \left(A_{c_y}^{gtbb} - A_{c_y}^{pbb}\right)^2} \quad (6)$$

$$c_h = \max\left(A_{c_y}^{gtbb}, A_{c_y}^{pbb}\right) - \min\left(A_{c_y}^{gtbb}, A_{c_y}^{pbb}\right) \quad (7)$$

Where, $c_h$ is the difference in height between the center points of the ground-truth box and the predicted box, $\sigma$ is the distance between the center points of the ground-truth box and the predicted box, and $(A_{cx}^{gtbb}, A_{cy}^{gtbb})$ represent the center coordinates of the ground-truth box and $(A_{cx}^{pbb}, A_{cy}^{pbb})$ represent the center coordinates of the predicted box.

The distance cost is used to calculate the center position and size of the detection box. The distance loss can ensure that the position and size of the detection box are correctly predicted. Its definition is as follows in Fig. 3.

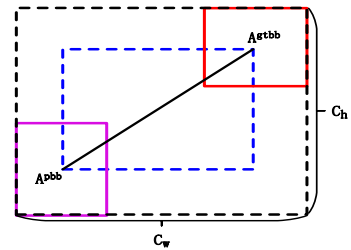

**Fig. 3** Detect distance loss of box center position and size

$$\Delta = \sum_{t=x,y} \left(1 - e^{-\gamma \rho_t}\right) = 2 - e^{-\gamma \rho_x} - e^{-\gamma \rho_y} \quad (8)$$

$$\rho_x = \left(\frac{A_{c_x}^{gtbb} - A_{c_x}^{pbb}}{c_w}\right)^2 \tag{9}$$

$$\rho_y = \left(\frac{A_{c_y}^{gtbb} - A_{c_y}^{pbb}}{c_h}\right)^2 \tag{10}$$

$$\gamma = 2 - \wedge \tag{11}$$

The shape cost is used to calculate the shape information of the detection box. For rotated boxes, the shape loss can ensure that the rotation angle, height, and width of the rotated box are correctly predicted. Its definition is as follows in Fig.4.

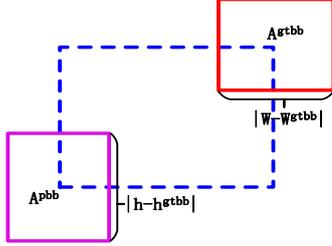

**Fig. 4** Schematic of the relation of IoU component contribution

$$\Omega = \sum_{t=w,h} \left(1 - e^{-\omega_t}\right)^\theta \tag{12}$$

$$\omega_w = \frac{|w - w^{gtbb}|}{max(w, w^{gtbb})} \tag{13}$$

$$\omega_h = \frac{|h - h^{gtbb}|}{max(h, h^{gtbb})} \tag{14}$$

The value of θ defines the shape cost for each data set and its value is unique. θ = 4 is calculated by the genetic algorithm in the paper.

IoU is the most commonly used loss function for target detection, which represents the intersection and union ratio of the real box and the prediction box. However, the IoU metric is used in the calculation, and the negative IoU is used for optimization during training and the negative IoU is defined as follows:

$$negative\ IoU = \frac{(MP)\inf}{(KP)\inf} - IoU(P, G) \tag{15}$$

Where *P* represents the predicted bounding box, and *G* represents the ground-truth bounding box.

To intuitively understand the relationship between the real box, the prediction box, and the IoU, the relationship between them is visualized as shown in Fig. 5.

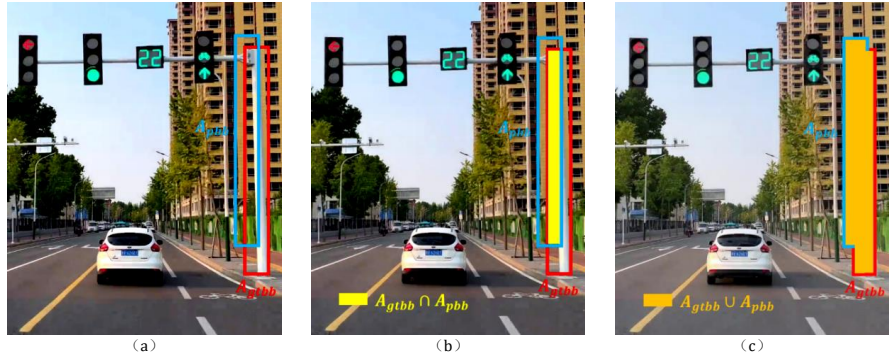

**Fig. 5** Visualization of IoU Loss. (a) the ground-truth bounding box (*A_gttb*) and the predicted bounding box (*A_pbb*); (b) the intersection of *A_gttb* and *A_pbb*; (c) the union of *A_gttb* and *A_pbb*

$$IoU = \frac{A_{gtbb} \cap A_{pbb}}{A_{gtbb} \cup A_{pbb}} \tag{16}$$

The IoU cost is used to calculate the overlapping part of the detection box. The IoU loss can ensure that the prediction of the detection box overlaps with the ground-truth object as much as possible, thereby improving the accuracy of detection. Its expression is as follows:

$$L_{IoU\ Cost} = 1 - IoU \tag{17}$$

The final expression of the SIoU loss function is as follows:

$$Loss_{MKS} = negativeIoU \cdot L_{IoU\ Cost} + \frac{\Delta + \Omega}{2} \tag{18}$$

### 2.2. Mixed domain attention model

The attention mechanism is a technique that simulates human attention and has been widely used in the field of deep learning. At present, there are two main types of attention mechanisms for processing feature maps: channel attention mechanism and spatial attention mechanism.

In target detection, with the continuous exploration of the model, it is difficult for the low-resolution basic feature map target detector to filter out important feature information from a large number of information channels. The channel attention mechanism, on the other hand, effectively solves this problem by allowing the model to focus on useful channel feature information, thus improving the performance of the model. Specifically, the spatial attention mechanism employs a set of convolutional kernels to convolve the feature map and learn the weight coefficients for each position. Specifically, given a feature with dimensions of H×W×C, a maximum pooling and average pooling operation are performed along the channel dimension to obtain channel descriptors of size 2×H×W×1. Then, a 7×7 convolutional layer and a Sigmoid activation function are applied to obtain the weight coefficients. Finally, these weight coefficients are used to compute a weighted sum of features for each position, generating a weighted spatial feature map. The expression for this process is as follows:

$$M_s(X) = \sigma\left(F_{conv}^{7\times7}\left(\left[X_{avg}^S, X_{max}^S\right]\right)\right) \tag{19}$$

Where, $X_{avg}^S$ and $X_{max}^S$ represent the feature map after global average pooling and global maximum pooling, $F_{conv}^{7\times7}$ represents the convolution operation with 7 × 7 kernel size, and σ is the sigmoid function.

According to the spatial attention weight map, the output of SAM could be expressed as:

$$\beta = F_{scale}(X, S) = X \otimes M \tag{20}$$

Where, $F_{scale}$ represents the spatial attention weighted mapping function, ⊗ represents the weighted multiplication.

Fusing the two mechanisms not only improves the performance of the model but also reduces the computational complexity and enhances the generalization ability. Currently, in deep learning, there are two main ways to connect the two attention mechanisms: cascading and parallel methods. The parallel attention mechanism [41] has a relatively simple structure, which may not fully utilize the correlations between different feature subspaces, leading to insufficient feature expression capabilities. Therefore, more layers need to be stacked. However, the cascade attention mechanism [42] can use the outputs of multiple attention modules for cascading, thereby better capturing the spatial and channel correlations of features and enhancing the feature expression capabilities. In addition, the cascade structure can also perform attention fusion on features maps of different scales, better adapting to objects of different scales, and its cascading structure is shown in Fig. 6.

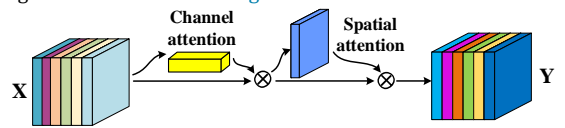

**Fig. 6** Cascade hybrid attention mechanism structure

## 2.3. Multi-scale feature pyramid model

The traditional feature pyramid structure (e.g., FPN[43], PAN) has the problems of high computational cost and insufficient information transmission. Therefore, this paper introduces the multi-scale feature fusion method RepGFPN [44]. However, this method has some defects in model performance, such as high computational cost, large memory consumption, gradient disappearance, etc. Therefore, this paper abandons ConvBNAct [45] and ConvWrapper and replaces them with convolutional layers to simplify the model structure, improve training speed and reduce model complexity. Then, by replacing BepC3 [46] with a lightweight convolutional layer, the model complexity is reduced and the training speed is improved. This makes the model more suitable for deployment in advanced driver assistance systems and even autonomous driving systems. Finally, a hybrid attention mechanism is introduced into the neck network of RepGFPN to enhance attention to important features and suppress unimportant feature information, thereby enhancing the accuracy and robustness of the model.

In summary, this paper embeds an optimized multi-scale feature-efficient fusion network architecture (MFFNA). The network architecture is shown in Fig. 7.

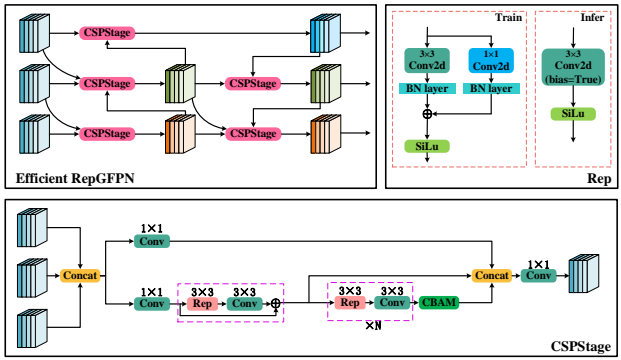

**Fig. 7** Multi-scale feature fusion network architecture.

According to the network architecture diagram, it can be seen that in the feature fusion module, this paper first uses the improved CSPStage [47] for convolutional fusion, and then performs structural reparameterization, specifically Conv2d and BN fusion. Therefore, the expression for BN in the ith channel of the feature map is as follows:

$$y_i = \frac{x_i - \mu_i}{\sqrt{\sigma_i^2 + \epsilon}} \gamma_i + \beta_i \quad (21)$$

Where the $\mu$ represents the mean, $\sigma^2$ represents the variance, $\gamma$ represents the weight, $\beta$ represents the bias, and $\epsilon$ is a very small constant that serves the purpose of preventing the denominator in the formula from being zero.

## 3. Experimental results and analysis

To evaluate the comprehensive performance of the proposed method, various experiments are conducted on the homemade dataset PSO 20023 and dataset VOC 2007. Next, the datasets and performance evaluation criteria are first introduced, then the experimental results are reported. Finally, the experimental results are analyzed.

### 3.1. Datasets

The home-grown datasets PSO 2023 and the public datasets VOC 2007 used in the experiments are as follows:
(1) PSO 2023: It contains 7 categories with a total of 6300 pictures. Moreover, the training and test sets were divided according to 9:1, with 5670 training sets and 630 test sets.
(2) VOC 2007: It contains 20 categories with a total of 5010 pictures. Moreover, the training and test sets were divided according to 9:1, with 4509 training sets and 501 test sets.

Validation of the improved method using the PSO 2023 homebrew datasets and the VOC 2007 public datasets. Meanwhile, to comprehensively evaluate the performance of the trained model, the model is tested on real roads in this paper. The experiments are mainly conducted in four complex environmental scenarios: rain and snow, a bright light environment, and a low light night.

For the PSO 2023 dataset details are given below: Currently, there is a lack of datasets for pole-like obstacles. Therefore, we performed a homemade collection and named it PSO 2023. we collected pole-like obstacles from real scenes in four seasons: spring, summer, autumn, and winter. The main idea is to collect at different times of the day, including morning, noon, afternoon, and evening. To enhance the generalization capability and robustness of the model, an additional 2800 images were collected for the homemade dataset under complex environments such as backlight, shade, cloudy, rainy, and snowy days, totaling 6300 images. Some of the images under various complex road environments are shown in Fig 8.

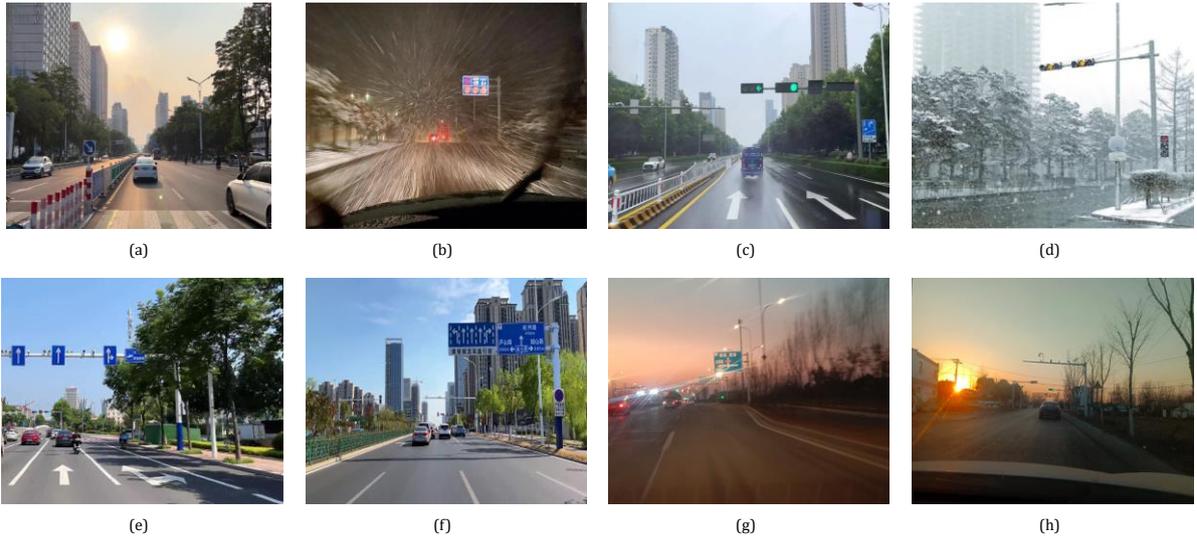

**Fig. 8** Images in partially complex road environments. (a) Afternoon glare scene; (b) Snowing scene at night; (c) Rained scene; (d) Snowed scene;(e) Foliage shaded environment; (f) Mutually obscured scenes; (g) Light pollution scenes at night; (h) Strong backlight environment in the evening.

### 3.2. Evaluation criteria

Precision, recall, AP, and mAP[48] are applied to evaluate the strengths and weaknesses of the model, and the formulas are shown below.

$$P_{precision} = \frac{TP}{TP + FP} \quad (22)$$

$$R_{recall} = \frac{TP}{TP + FN} \quad (23)$$

$$AP = \int_0^1 P(R)dR \quad (24)$$

$$mAP = \frac{1}{N} \sum_{n \in N} AP(n) \quad (25)$$

where TP denotes the number of correctly detected targets, FP denotes the number of incorrectly detected targets, FN denotes the number of undetected targets, and N denotes the number of classes that need to be classified in total. Recall indicates the number of correctly predicted samples as a percentage of all samples that are positive cases. mAP can be used as a comprehensive evaluation metric for individual category detection. higher AP values indicate better detection of a category, and mAP is a comprehensive evaluation of the entire network.

### 3.3. Parameter settings

The following settings are made when the model is trained. This experiment uses the optimal training of the model using the SGD optimizer with an input image size of 640×640, an initial learning rate of 1e-2, and several iterations of 300, the batch size is 16. The IoU threshold and momentum are set to 0.2 and 0.937. The freeze part batch size is 8, and freeze training is for 20 cycles. At the same time, the hardware configuration of the experiments is as follows: the experimental platform is Windows 10, the processor is Intel Core i9-12900H, equipped with NVIDIA GeForce RTX3090-24GB, the development environment is Pycharm2022, Python3.9, the deep learning framework is Pytorch1.13.1, using CUDA11.6.0 and CUDNN8.6.0 for image acceleration.

### 3.4. Model Training

Using the above parameter settings to train the models for Faster-RCNN, YOLOv5s, and the improved algorithm, respectively, the final total loss convergence curves and mAP comparison plots are shown in Fig. 9(a) and 9(b). From Fig. 9, it is easy to see that the MFMAM-YOLOv5s target detection method proposed in this paper converges faster than the remaining two algorithms. Meanwhile, all three algorithms rise rapidly in the first 25 rounds. However, after 25 rounds, the rising trend of the curve of the algorithm proposed in this paper is more obvious. The experiments show that the method not only converges easily but also has high stability and accuracy.

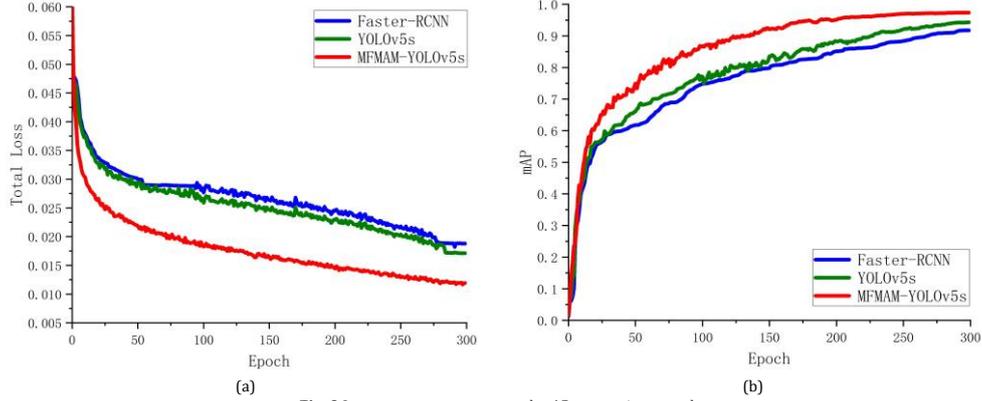

**Fig. 9** Loss convergence curve and mAP comparison graph

#### 3.4.1. Ablation experiment

In this paper, to verify the effectiveness of the proposed multiscale feature mixture attention algorithm, ablation experiments were conducted on the MFMAM-YOLOv5s algorithm on the homemade PSO 2023 datasets. The experimental results are shown in Table 1. It can be seen from Table 1 that the mAP value increased from 94.3% to 95.6% after using the improved loss function MKS in the original algorithm. Meanwhile, the number of floating-point operations is reduced from 16.0 G to 15.8 G. The detection rate of the model is increased from 303.0 f/s to 454.5 f/s, and the parameters are also reduced by 0.01 M. Therefore, the improved loss function has good results for pole-like obstacle detection in complex environments, both in terms of detection accuracy and real-time performance.

Based on the MKS loss function, the hybrid attention mechanism CBAM is embedded into the algorithm. The mAP value is 1.9 % higher than the original algorithm. The number of floating-point operations only increased by 0.6 G, and the parameters only increased by 0.51 M. Therefore, the hybrid attention mechanism CBAM is introduced into the algorithm to focus on important features and suppress non-important information, which is very effective for pole-like obstacle detection in complex environments.

Similarly, an improved multi-scale feature extraction module is embedded based on the MKS loss function and mixed attention mechanism. The mAP value is 3.1 % higher than that of the original algorithm, and the parameters and floating-point operations of the model are only increased by 2.74 M and 4.3 G. Although the number of floating-point operations and parameters are slightly increased, the detection accuracy is met while ensuring real-time performance. It can be seen that the algorithm proposed in this paper gives full play to the accuracy and real-time performance in the detection of pole-like obstacles in complex environments.

#### 3.4.2. Comparison of Various Loss Functions

To verify the effects of GIoU, DIoU, CIoU, and the improved loss function on the accuracy and real-time performance of pole-like obstacle detection, the following four sets of experiments were conducted in this paper. The experimental results are shown in Table 2. From this table, it can be seen that the mAP_0.5 of the improved loss function is 1.4%, 1.5%, and 1.3% higher than GIoU, DIoU, and CIoU, respectively. Similarly, the FPS increases by 221.9 f/s, 227.2 f/s, and 232.3 f/s, respectively. volume and FLOPS decrease by 0.6 MB and 0.2 G, respectively, while the parameters are reduced to 7.03 M.

**Table 1**
Ablation experiment of MFMAM-YOLOv5s.

| MKS | CBAM | Multi-scale | mAP/% | FLOPs/G | FPS/f/s | Volume/MB |
|---|---|---|---|---|---|---|
| × | × | × | 94.3 | 16.0 | 303.0 | 14.40 |
| √ | × | × | 95.6 | 15.8 | 454.5 | 13.77 |
| √ | √ | × | 96.2 | 16.6 | 454.5 | 14.78 |
| √ | √ | √ | 97.4 | 20.3 | 400.0 | 19.04 |

**Table 2**
Comparison results of different loss functions.

| Method | Parameters/M | mAP_0.5/% | FLOPs/G | FPS/f/s | Volume/MB |
|---|---|---|---|---|---|
| GIoU-YOLOv5s | 7.55 | 94.2 | 16.0 | 232.6 | 14.4 |
| DIoU-YOLOv5s | 7.04 | 94.1 | 16.0 | 227.3 | 14.4 |
| CIoU-YOLOv5s | 7.04 | 94.3 | 16.0 | 222.2 | 14.4 |
| Improve YOLOv5s | 7.03 | 95.6 | 15.8 | 454.5 | 13.8 |

### 3.5 Comparison of different algorithms

To reflect the effect of the algorithm improvement, this paper conducted experimental validation using the target detection algorithms YOLOv4, YOLOv4-tiny, Faster-RCNN, and YOLOv5 on the homemade dataset PSO 2023 and the public dataset VOC 2007, respectively. The experimental results are shown in Tables 3 and 4. From the tables, it is easy to find that the two-stage Faster-RCNN do not bring higher accuracy, but rather slower detection frame rates, even though they use the Resnet50 backbone network. For YOLOv4 although the detection accuracy is relatively good, the model has difficulties in deployment. YOLOv4-tiny and Yolov5s, as commonly used lightweight algorithms, will have some advantages in the deployment of the model, but it is difficult to satisfy the complex environment of the assisted driving perception technology due to the lack of prominence in detection accuracy and real-time performance. Compared with the models proposed in this paper, MFMAM-YOLOv5s, MFMAM-YOLOv5m, MFMAM-YOLOv5l, and MFMAM-YOLOv5x have different degrees of improvement in mAP. Especially, MFMAM-YOLOv5s has a significant improvement in both FPS and mAP over YOLOv5s. Compared to the above models, the MFMAM-YOLOv5s model achieves a certain balance between speed mixed with accuracy, which is more suitable for the deployment of assisted driving systems.

**Table 3**
Performance of different algorithms on the PSO 2023 datasets.

| Model | Backbone | mAP/% | FPS/f/s | Volume/MB |
|---|---|---|---|---|
| YOLOv4 | CSPDarknet53 | 91.6 | 36.9 | 244.5 |
| YOLOv4-tiny | CSPDarknet53-Tiny | 83.9 | 153.9 | 22.5 |
| Faster-RCNN | Resnet50 | 82.3 | 23.0 | 108.0 |
| Faster-RCNN | VGG16 | 88.5 | 17.6 | 521.0 |
| YOLOv5s | CSPDarknet53 | 94.3 | 303.0 | 14.4 |
| YOLOv5m | CSPDarknet53 | 94.6 | 263.2 | 40.3 |
| YOLOv5l | CSPDarknet53 | 95.3 | 156.3 | 88.6 |
| YOLOv5x | CSPDarknet53 | 95.7 | 94.3 | 165.2 |
| MFMAM-YOLOv5s(our) | CSPDarknet53 | 97.4 | 400.0 | 19.0 |
| MFMAM-YOLOv5m(our) | CSPDarknet53 | 97.8 | 277.8 | 61.3 |
| MFMAM-YOLOv5l(our) | CSPDarknet53 | 98.2 | 147.1 | 97.1 |
| MFMAM-YOLOv5x(our) | CSPDarknet53 | 98.8 | 88.5 | 179.7 |

**Table 4**
Performance of different algorithms on the VOC 2007 datasets.

| Model | Backbone | mAP/% | FPS/f/s | Volume/MB |
|---|---|---|---|---|
| YOLOv4 | CSPDarknet53 | 82.3 | 22.4 | 231.3 |
| YOLOv4-tiny | CSPDarknet53-Tiny | 77.6 | 108.2 | 21.6 |
| Faster-RCNN | Resnet50 | 78.5 | 13.0 | 112.0 |
| Faster-RCNN | VGG16 | 86.7 | 7.4 | 523.1 |
| YOLOv5s | CSPDarknet53 | 80.2 | 61.0 | 13.8 |
| YOLOv5m | CSPDarknet53 | 86.4 | 21.3 | 80.5 |
| YOLOv5l | CSPDarknet53 | 90.2 | 12.1 | 177.2 |
| YOLOv5x | CSPDarknet53 | 91.8 | 6.0 | 321.3 |
| MFMAM-YOLOv5s(our) | CSPDarknet53 | 84.6 | 131.6 | 27.3 |
| MFMAM-YOLOv5m(our) | CSPDarknet53 | 88.0 | 81.3 | 61.3 |
| MFMAM-YOLOv5l(our) | CSPDarknet53 | 91.3 | 66.7 | 185.0 |
| MFMAM-YOLOv5x(our) | CSPDarknet53 | 92.5 | 33.1 | 378.7 |

To demonstrate the detection effect of the improved algorithms more intuitively, this paper demonstrates the detection of different algorithms on the homemade dataset PSO 2023, respectively, as shown in Fig. 10.

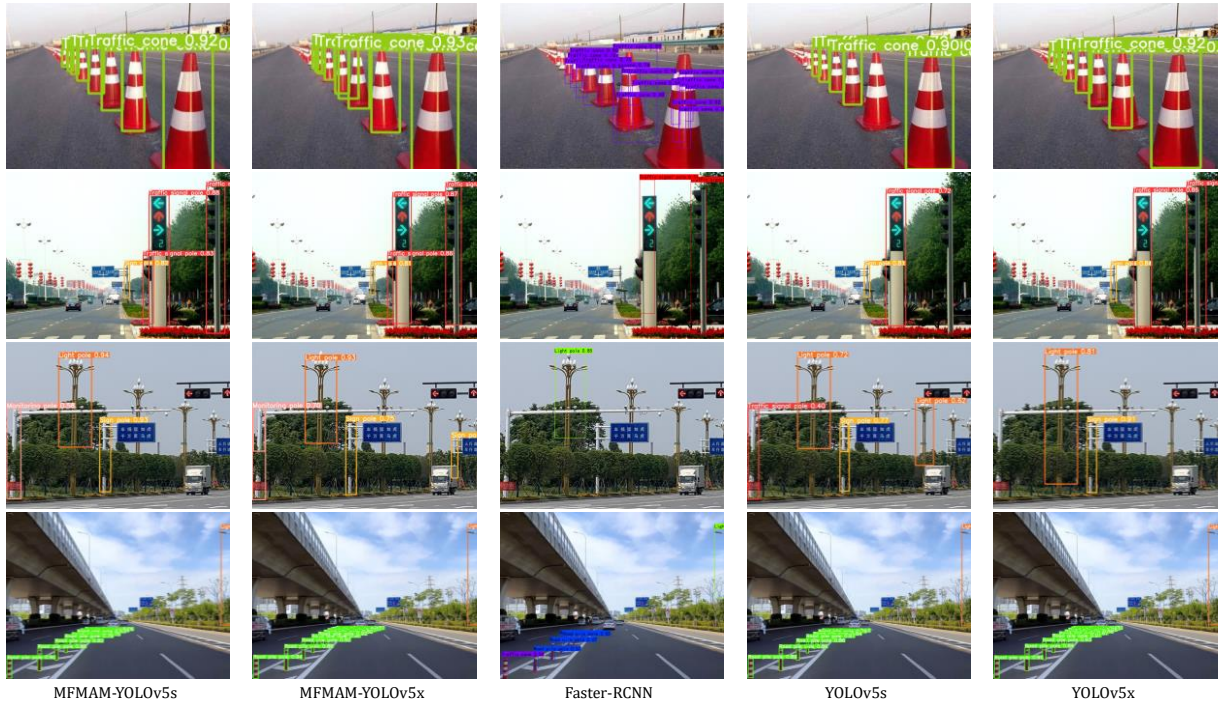

| MFMAM-YOLOv5s | MFMAM-YOLOv5x | Faster-RCNN | YOLOv5s | YOLOv5x |

**Fig.10** Comparison of test results for different algorithms in homemade datasets

It can be seen from the figure that the MFMAM-YOLOv5s and MFMAM-YOLOv5x algorithms not only have more targets detected than the YOLOv5s and YOLOv5x algorithms but also have higher confidence in detecting targets. At the same time, the false detection rate and missed detection rate of the proposed method are significantly lower than the original algorithm during testing. By comparing with the two-stage object detection algorithm Faster-RCNN, it is found that the YOLOv5 and MFMAM-YOLOv5 algorithms have higher accuracy than Faster-RCNN in both small-scale target detection and large-scale object detection. This shows that our proposed method improves the overall performance of the model on both large and small models. Therefore, MFMAM-YOLOv5 not only enriches the deep semantic information of the feature map but also improves the adaptability of the network to targets of different sizes.

### 3.5.1. Detection Comparison Diagram

Since the performance of the trained model from the Faster-RCNN algorithm differs significantly from the remaining two algorithms. Therefore, in this paper, the performance of the original algorithm YOLOv5s and MFMAM-YOLOv5s is verified in real-time on the homemade dataset PSO 2023 and real roads. The following is the comparison and analysis of some pole-like obstacle detection sample images in complex environments randomly selected from the detection results of the two algorithms. First, the test is performed in a rainy environment [49] as shown in Fig. 11.

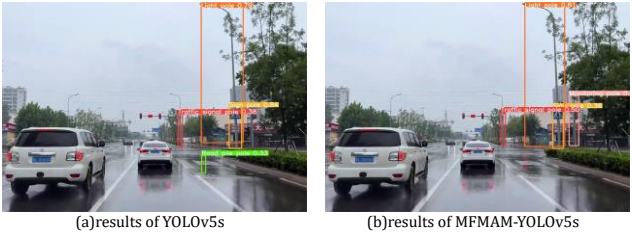

(a)results of YOLOv5s　　　　(b)results of MFMAM-YOLOv5s

**Fig. 11** Comparison diagram of pole-like obstacles detection in rainy weather

From the experimental comparison of Fig. 11 (a) and Fig. 11 (b), it can be seen that red and green light reflection is easy to occur in rainy weather, which leads to the false detection of the original algorithm. However, the MFMAM-YOLOv5s algorithm proposed in this paper shows strong anti-interference ability in rainy and complex environments. By suppressing the feature information of interference, the robustness and generalization ability of the model is continuously improved, and the accurate detection of pole-like obstacles in complex environments is realized. Pole-like obstacle detection in snow and ice weather [49] is shown in Fig. 12.

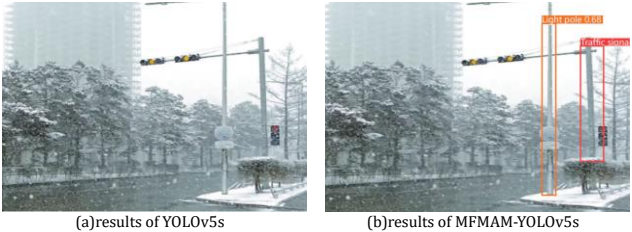

(a)results of YOLOv5s　　　　(b)results of MFMAM-YOLOv5s

**Fig. 12** Comparison diagram of pole-like obstacle detection in ice and snow weather

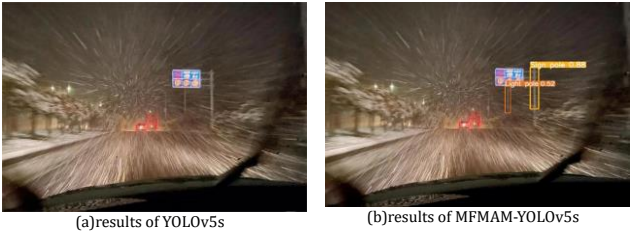

(a)results of YOLOv5s　　　　(b)results of MFMAM-YOLOv5s

**Fig. 13** Comparison of pole-like obstacle detection in dark snow and ice weather

From Fig. 12 and Fig.13, it is found that the original algorithm is prone to missed detection of pole-like obstacle detection in snowy weather. The algorithm proposed in this paper has good anti-interference performance and can complete the detection of pole-like obstacles in complex environments, and has achieved very effective results. The experimental results under the condition of blade occlusion and pole-like obstacle overlap are shown in Fig.14 and Fig15.

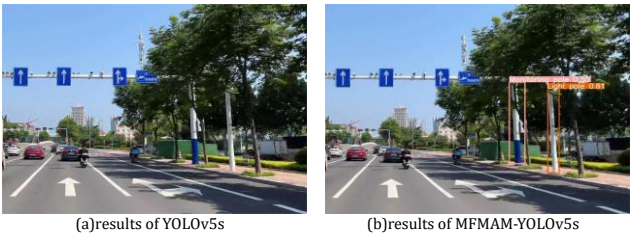

(a)results of YOLOv5s　　　　(b)results of MFMAM-YOLOv5s

**Fig. 14** Comparison diagram of pole-like obstacle detection with leaf occlusion

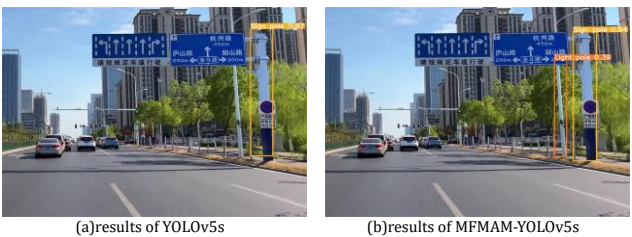

(a)results of YOLOv5s　　　　(b)results of MFMAM-YOLOv5s

**Fig. 15** Detection comparison diagram of pole-like obstacle overlapping environment

It can be seen from Fig. 14 and Fig. 15 that because some features are occluded, the original algorithm does not extract key feature information, resulting in missed detection. The MFMAM-YOLOv5s algorithm suppresses the occlusion feature information according to the hybrid attention mechanism, and focuses on the feature information of the pole-like obstacle feature space, thereby improving the robustness and detection accuracy of the model. To better reflect the detection performance of the model in darker light, the experimental results are shown in Fig. 16. In the dark environment, the original algorithm is prone to false detection and missed detection, while the MFMAM-YOLOv5s algorithm completes the detection of pole-like obstacles well.

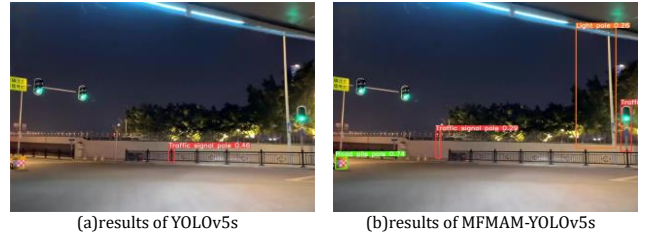

(a)results of YOLOv5s　　　　(b)results of MFMAM-YOLOv5s

**Fig. 16** Detection comparison diagram in a dark environment

Based on the above analysis, the MFMAM-YOLOv5s method proposed in this paper can well complete the detection of pole-like obstacles in complex environments. According to the characteristics of the MKS loss function, the problem of overlapping multiple prediction boxes and optimal matching is solved. Then, combined with the multi-scale feature extraction module, the feature space information is extracted and passed to the mixed attention mechanism. Therefore, the interference feature information is suppressed, the attention to the key feature information of the pole-like obstacle is improved, and the accurate detection of the pole-like obstacle in the complex environment is completed.

### 3.6. The effect of light on detection

In complex environments, it is important to consider not only the effect of light intensity but also the noise problem due to rain and snow. As shown in Fig. 17, the image is first masked as a control to analyze the pole obstacle region with different brightness gray values.

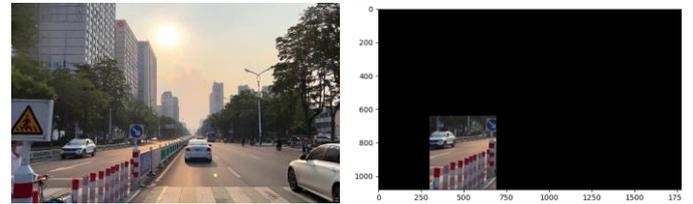

**Fig. 17** Mask processing analysis chart

#### 3.6.1. Algorithm Robustness Verification

Since assisted driving requires high real-time accuracy, the following experiment only needs to compare and analyze the original algorithm and the algorithm in this paper. Therefore, to further verify the robustness of the proposed method, The original algorithm and the algorithm in this paper are analyzed for the detection of targets with different bright gray values and peak signal-to-noise ratios, respectively. First, 0-255 is divided and tested for every 10 gray values, and then the results of missed detection and detection failure have experimented again with 1 gray value until the limit position of missed detection and detection failure is determined. Considering the practical application of the onboard camera, this paper removes the gray value of the extreme black-and-white stage before experimenting.

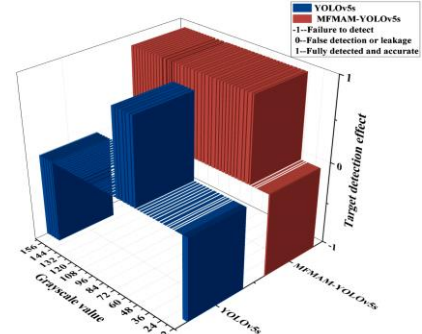

**Fig. 18** Effect on detection at different gray values

As shown in Fig 18, the original algorithm fails to detect between 12-15 and 150-164 gray values; it can detect between 16-58 and 80-149 gray values but there is a leakage; however, it can detect between 59-79 gray values completely and accurately. The robust performance of this paper's algorithm is shown in Fig. 18, the detection failure at 12 gray values; between 13-28 and 149-164 gray values can be detected but there is a leakage; however, between 29-148 gray values can be detected completely not and accurate. From the above experimental comparison results, it can be seen that the robustness of the proposed algorithm is stronger than the original algorithm.

### 3.6.2. Verification of the universality of the MFMAM-YOLOv5 algorithm

Considering the practical application of the onboard camera, this paper removes the gray values of the extreme black-and-white stages. As shown in Fig. 19, firstly, the gray values between 18-160 are divided into every 10 gray values, totaling 15 images with different brightness; then the 15 images are detected and masked respectively; finally, the luminance histograms are analyzed for the pole obstacles in the masked control region.

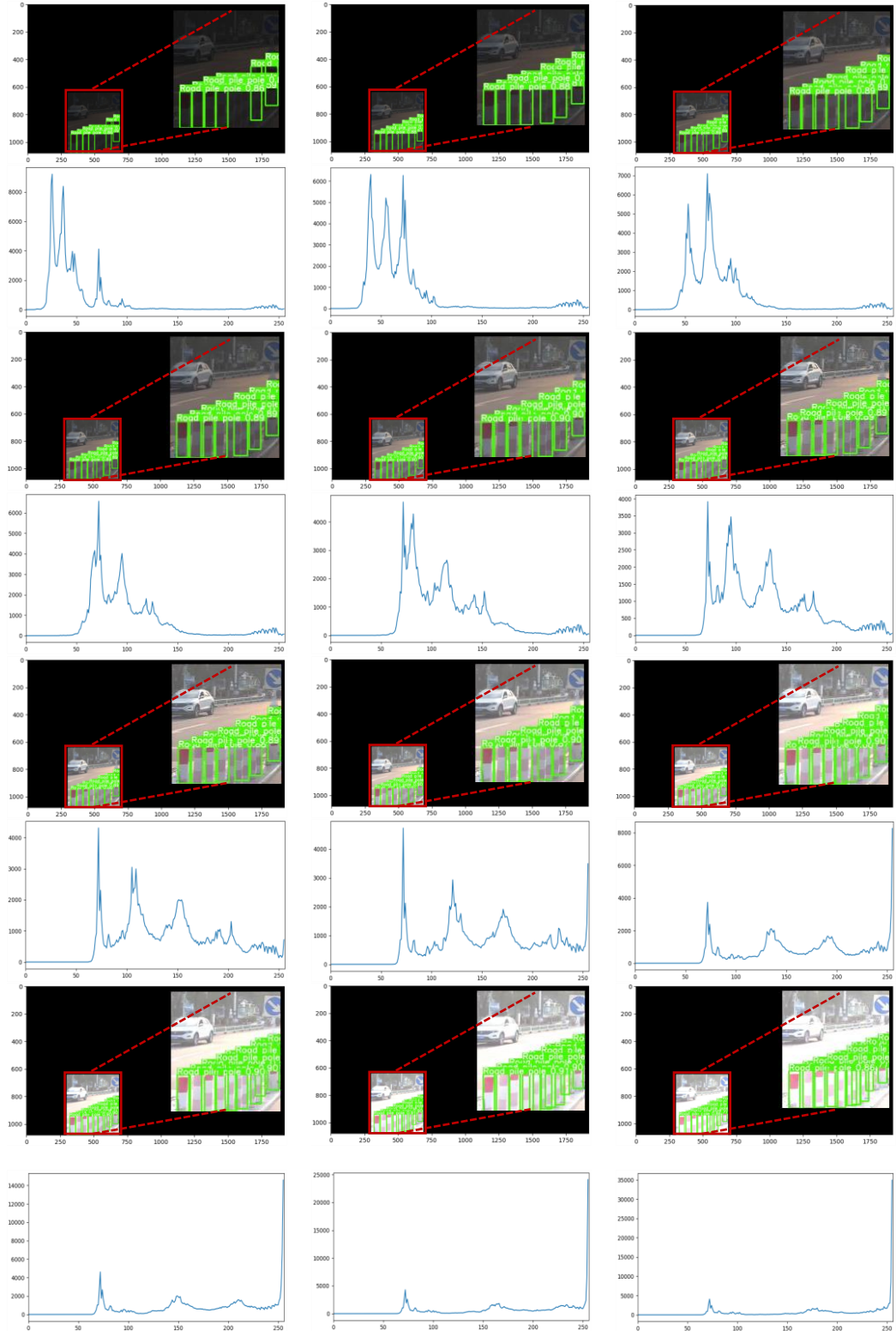

**Fig. 19** Histogram analysis of object detection at different grayscale values

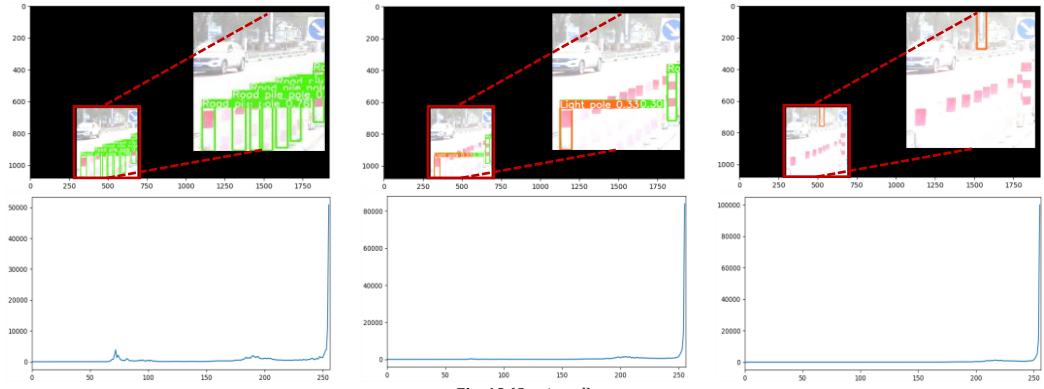

**Fig. 19** (Continued)

Through the experiment, it is easy to find that the cases of omission, misdetection, and detection failure for the detection of pole obstacles are concentrated between 19-28 and 149-160, i.e., the wave peaks are concentrated on the two sides of the histogram; while the cases of error-free detection are concentrated between 29-148, i.e., the wave peaks are concentrated in the middle of the histogram.

### 3.6.3 The effect of noise on detection

In the rain and snow weather not only to consider the problem of light intensity, but also to consider the noise problem of rain and snow. Therefore, in this paper, the original algorithm and the improved algorithm are used to detect and analyze the targets with noise strength respectively. Firstly, the images of rain and snow weather are selected and incrementally detected for mean and var from 0-0.1 in units of 0.001. Then the experiment is repeated with the results of missed detection and detection failure in units of 0.0001 until the limit positions of missed detection and detection failure are determined. As shown in Fig. 20 the original algorithm fails to detect between 0-18.77 DB peak SNR[50]; misses between 18.78-25.68 DB peak SNR; and detects without error between 25.69-31.97 DB peak SNR. However, for the robust performance of the algorithm in this paper it can be seen in Fig.20 that there is a failure of detection between 0-14.70DB peak SNR; there is a miss detection between 14.71-19.70DB peak SNR; No detection failure between 19.76-31.97DB peak SNR ratio. From the above experimental comparison results, it can be seen that the robustness of the algorithm proposed in this paper is stronger compared to the original algorithm.

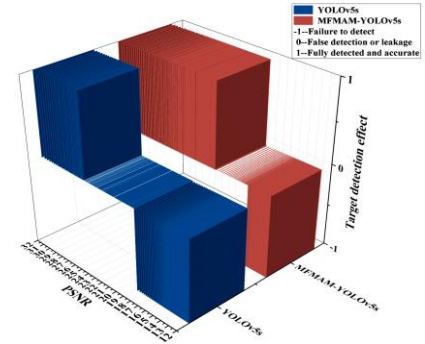

**Fig. 20** Effect on detection at different peak SNR ratios

To further verify the universality of the algorithm under noise, the initial position and limit position of accurate detection is first determined, and then two sets of images are randomly selected for testing. Similarly, the same method is used for missed detections, false detections, and detection failures. As shown in Fig. 21, two sets of renderings are shown for the initial position, and the limit position, and randomly selected in between.

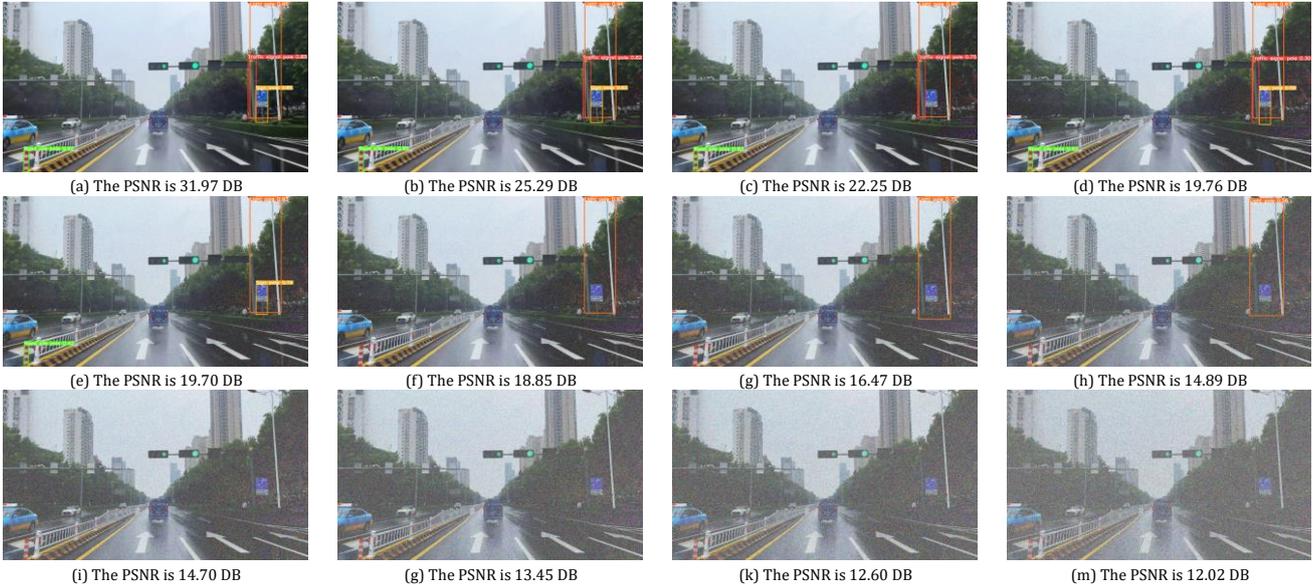

**Fig. 21** Object detection effect diagram under PSNR

Among them, (a)-(d) belongs to the accurate detection effect diagram; (e)-(h) is the missed detection and false detection; (i)-(m) is the detection failure effect diagram. Through the above experimental analysis, it can be seen that the universality of the algorithm in light-intensity scenes and random noise scenes is stronger than that of the original algorithm. In summary, the algorithm not only has good applicability to rod obstacle detection but also has a good detection effect on public datasets.

### 4. Conclusions

This paper proposes a multi-scale mixing attention mechanism approach for pole-like obstacle detection in complex environments. The algorithm mainly includes the MKS loss function module, multi-scale feature extraction module, and mixed attention mechanism module. Firstly, the MK function in the MKS loss function enables the model to pay more attention to the spatial relationship between objects and optimize the loss function. Because the multiple prediction boxes of the original model overlap and wander back and forth, the inference time and detection speed become slower. The improved MKS loss function determines each detection box and its most matched real box by calculating the minimum cost matching, to solve the problem that multiple detection boxes overlap with the same real box.

Therefore, the inference time and detection speed of the model are greatly improved. Then, based on the characteristics of the multi-scale feature extraction module, the features processed by the MKS loss function are sent to the channel for segmentation, and at the same time, the information of multi-scale feature extraction is collected by convolution of different sizes. Finally, the collected feature information is sent to the mixed attention mechanism for multi-scale feature learning. According to the characteristics of the channel attention mechanism and spatial attention mechanism, the interference information in complex environments is suppressed, focusing on the key feature information of pole-like obstacles. The proposed network not only achieves good performance on the self-made dataset PSO 2023 but also works effectively on the public dataset VOC 2007. At the same time, through experimental analysis on the self-made dataset PSO 2023, it is found that the proposed algorithm not only increases the mAP value by 3.1% compared with the original algorithm but also increases the detection rate from 303.0 f/s to 400.0 f/s. While accurately detecting rod-shaped obstacles, real-time detection is also guaranteed. Therefore, in complex environments, the overall performance of the MFMAM-YOLOv5s algorithm is better than the original algorithm.

### CRediT authorship contribution statement

**Lei Cai:** Methodology, Investigation, Writing – original draft Visualization. **Hao Wang:** Writing – review & editing. **Congling Zhou:** Funding acquisition, Supervision. **Yongqiang Wang:** Formal analysis, Resources. **Boyu Liu:** Software, Validation.

### Data availability

Publicly available datasets were analyzed in this study. This data can be found here: https://pjreddie.com/projects/pascalvoc-dataset-mirror/.

### Declaration of Competing Interest

The authors declare that they have no known competing financial interests or personal relationships that could have appeared to influence the work reported in this paper.

### Acknowledgements

This work was supported by the Beijing Zhongke Huiyan Technology Co., Ltd. (No.2200010047).

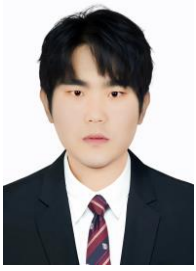

**Lei Cai** received the B.Eng. degree from Jiangxi University of Technology, College of Applied Science in 2020. He is currently pursuing his master's degree at Tianjin University of Science and Technology. His current research interests are visual detection and its application in environment perception.

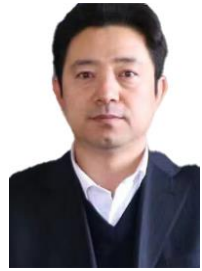

**Yongqiang Wang** received the B.Eng. and M.S. degrees from Beijing University of Aeronautics and Astronautics in 1986 and 1989, respectively. He is currently a professor in the School of Mechanical Engineering at Tianjin University of Science and Technology. He has published more than 100 academic papers in China and abroad. His current research interests include machine vision inspection technology and applications, mechatronics product development and mechanical engineering testing technology.

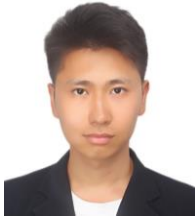

**Hao Wang** received the B.Eng., M.S. and Ph.D. degrees from China University of Petroleum in 2013, 2015 and 2021, respectively. He has published two SCI papers and one EI paper. Her current research interests include oil and gas pipeline detection technology and oil and gas well mechatronics integration system.

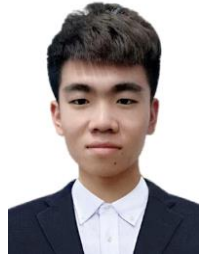

**Boyu Liu** received the B.Eng degree in mechanical design, manufacturing and automation from Shijiazhuang University of Railways in 2021. He is currently pursuing a master's degree in mechanical engineering at Tianjin University of Science and Technology. His main research interests are machine vision and mechanical parameter measurement and control.

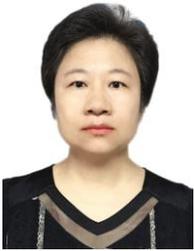

**Congling Zhou** received the B.Eng., M.S. and Ph.D. degrees from Tianjin Institute of Light Industry, Tianjin University of Science and Technology and Saga University in 1999, 2002 and 2005, respectively. She is currently an Associate Professor in the School of Mechanical Engineering at Tianjin University of Science and Technology. She has published more than 40 academic papers in China and abroad. Her current research interests include vision inspection techniques and applications as well as measurement methods and systems for mechanical parameters.